\documentclass[conference]{IEEEtran}
\IEEEoverridecommandlockouts
\usepackage{amsmath,amssymb,amsfonts}
\usepackage{algorithmic}
\usepackage{graphicx}
\usepackage{textcomp}
\usepackage{xcolor}
\usepackage{url}
\usepackage{svg}
\usepackage{multirow}
\usepackage{listings}
\definecolor{codegreen}{rgb}{0,0.6,0}
\definecolor{codegray}{rgb}{0.5,0.5,0.5}
\definecolor{backcolour}{rgb}{0.95,0.95,0.92}
\usepackage{cite}

\begin{document}

\title{TalkWithMachines: Enhancing Human-Robot Interaction for Interpretable Industrial Robotics Through Large/Vision Language Models\\
%{\footnotesize \textsuperscript{*}Note: Sub-titles are not captured for https://%ieeexplore.ieee.org  and should not be used}
%\thanks{Identify applicable funding agency here. If none, delete this.}
}
\author{\IEEEauthorblockN{Ammar N.~Abbas}
\IEEEauthorblockA{\textit{School of Computer Science} \\
\textit{Technological University Dublin}\\
Dublin, Ireland \\
ORCID: 0000-0002-2578-5137}
\and
\IEEEauthorblockN{Csaba Beleznai}
\IEEEauthorblockA{\textit{Assistive \& Autonomous Systems} \\
\textit{AIT Austrian Institute of Technology}\\
Vienna, Austria \\
ORCID: 0000-0003-1880-2979}
}

\maketitle

\begin{abstract}
TalkWithMachines aims to enhance human-robot interaction by contributing to interpretable industrial robotic systems, especially for safety-critical applications. The presented paper investigates recent advancements in Large Language Models (LLMs) and Vision Language Models (VLMs), in combination with robotic perception and control. This integration shall allow robots to understand and execute commands given in natural language and to perceive their environment through visual and/or descriptive inputs. Moreover, by translating the LLM's internal states and reasoning into text that humans can easily understand, it shall ensure that operators gain a clearer insight into the robot's current state and intentions, which is essential for effective and safe operation. Our paper outlines four LLM-assisted simulated robotic control workflows, which explore (i) low-level control, (ii) the generation of language-based feedback that describes the robot's internal states, (iii) the use of visual information as additional input, and (iv) the use of robot structure information for generating task plans and feedback, taking the robot’s physical capabilities and limitations into account. The proposed concepts are presented in a set of experiments, along with a brief discussion. Project description, videos and supplementary materials will be available on the project website: \url{https://talk-machines.github.io}.
\end{abstract}


\section{Introduction}
\label{sec:intro}
In recent years, research on autonomous systems has increasingly concentrated on aspects that ensure these systems are interpretable and easy to understand, particularly in sectors where safety and user trust are of primary importance \cite{kuznietsov2024explainable,he2021challenges}. This heightened emphasis on interpretability underscores the importance of creating natural forms of human-machine interaction, such as low-level control based on human language \cite{tang2023saytap,kwon2023language}. In safety-critical industries, however, this is often not recommended, and formal/traditional methods are used to develop verifiably safe autonomous systems \cite{yin2024formal,luckcuck2023using}. On the other hand, despite their existing shortcomings in providing deterministic and reliable outputs, recent breakthroughs in Large Language Models (LLMs) and Vision Language Models (VLMs) present promising possibilities as a potential intermediate communication and reasoning layer between humans and industrial machines or robots \cite{brohan2023can,jin2024robotgpt,lynch2023interactive,vemprala2024chatgpt}. These developments encourage us to investigate the capabilities of LLMs and VLMs for robotic control and improved interpretability, particularly in translating a machine's internal states, observations, predictions, and actions into language that is easily understandable by humans \cite{yang2021review,chen2023gap}.

We build on the recent advancements in LLMs and VLMs to demonstrate their potential for robotic manipulation and perception. Here, \textit{perception} refers to the text- and/or image-based representation of both the robot's internal and external states as input, used for reasoning. Much of the recent research has concentrated on utilizing LLMs in robotics as high-level planning modules \cite{huang2023inner,huang2024copa} or coding platforms \cite{liang2023code,singh2023progprompt,di2024keypoint}. Since high-level concepts are effectively represented in text while low-level signals and percepts often exist in different domains, it is commonly assumed that LLMs are not suited for low-level control \cite{tang2023saytap,kwon2023language}. Consequently, our objectives concentrate on two less-explored areas: (i) language-based low-level control, and (ii) transforming machine states and intentions into interpretable text, referred to as \textit{verbalization} of machine states. 

LLMs have proven to be considered as the general pattern machines \cite{mirchandani2023large}, capable of understanding and combining low-level patterns for complex control strategies \cite{tang2023saytap}. \cite{mirchandani2023large} demonstrates the LLMs as general pattern machines for robotic manipulation with examples involving sequence transformation, sequence completion, sequence improvement, and as a controller. \cite{tang2023saytap} similarly utilizes the foot contact pattern as an interface to bridge the communication between human language commands and low-level quadruped robot control. These combinations of basic control patterns demonstrate complex behavior understanding and lead to the communication of actions to the user and its perception of the environment, which is also the aim of this paper. We utilize robotic arm manipulation within a simulation environment as our experimental setup and propose an improved interface concept for communication between the user and the robot.

The questions that we aim to address through this study are: 
\\1. Can low-level control command patterns be generated through few-shot human language-to-pattern examples given to the LLM?
2. Are LLMs capable of generating complex trajectories (such as grasping an object while avoiding obstacles) using basic control patterns?
3. Can LLMs perceive and maintain contextual awareness of the environment from a single observation and prompt the user about anomalies (such as object penetration, unobserved conditions, environmental constraints, or indications of obstructions)?
4. Can LLMs make autonomous decisions based on the real-world properties of the environment (such as object characteristics or ground conditions)?     
5. Are LLMs in robotic manipulation tasks generalizable to incorporate indirect communication cues based on common sense? 
6. Can a stack of time-consecutive image frames depicting the environment enhance the situational awareness of the VLM?
7. Can the information from the Unified Robot Description Format (URDF) \cite{tola2024understanding} help LLM/VLM to become structure-aware and identify out-of-reach or singularity conditions?
8. Does integrating specific constraints within the environment help LLM perceive the environment for unsafe command execution? 
9. Can LLM extrapolate or predict the next steps in a control sequence or based on the environment perception?

The paper is structured as follows: Section \ref{sec:sota} provides an overview on related methods. Section \ref{sec:method} describes our human-robot interaction methodology via LLMs/VLMs. Section \ref{sec:doe} explains the design of experiments used for addressing the outlined research questions. Section \ref{sec:results} provides our experimental findings. Finally, Section \ref{sec:conclusion_future} concludes the paper and outlines future directions.

%===============================================================================
\section{State-of-the-Art}
\label{sec:sota}
LLMs and specifically Generative Pre-Trained Transformers (GPT) version 4 \cite{achiam2023gpt} have proven to generalize beyond natural language as zero-shot or few-shot models towards the domain of control and automation \cite{yang2024human,shentu2024llms}.
\\ \textbf{Large/Vision Language Models for Robotics:} \cite{hu2023toward} provides an extensive literature review on the application of foundation models towards general-purpose robotics, discussing the current gap and future directions. \cite{kira2022llmroboticspaperslist} present a broad review on LLMs for robotics. Common research categories within the domain of LLM-assisted robotics can be defined as (i) high-level planning, (ii) context/structure-aware perception, (iii) code generation, and (iv) low-level control. 
\\ \textbf{High-Level Planning:} \cite{Kambara2024jun} use a multimodal LLM approach as a planner to combine audio, video, speech, images, and text as input to generate action sequences. LLM-based planners are used by \cite{parakh2023lifelong} to overcome the limitation of acquiring new skills that are not available in the predefined set of skills, proposing a lifelong learning framework.
\\ \textbf{Context- and Structure-Aware Perception:} \cite{sharan2024plan} demonstrate a step-wise closed-loop planning and control architecture through visual feedback using a diffusion model. LLM is used to generate step-by-step textual subgoals which are then transformed into visual subgoals through a diffusion model for planning. A goal-conditioned policy then transforms these visual subgoals to control actions. Similarly, \cite{jin2023alphablock, song2023llm} employ the use of language and visual observation to form a closed-loop context-aware perceptive planner. \cite{hu2023look, shridhar2022cliport} include the attributes of objects and their spatial information, as well. 
\\ \textbf{Code Generation:} LLMs have been used to demonstrate the capabilities of generating robot manipulation plans and the corresponding codes to accomplish it \cite{toberg2023generation}. \cite{vemprala2024chatgpt} assess the effectiveness of ChatGPT by integrating a high-level function library on various techniques, including free-form dialogue, XML parsing, code synthesis, task-specific prompting, and closed-loop reasoning. These methods are tested across a spectrum of tasks, from simple logical problems to complex scenarios like aerial navigation and robotic manipulation.
%------------- CONTRIBUTION TABLE ------------------------------------------------------------
\begin{table}[!t]
    \centering
        \caption{Qualitative comparison to state-of-the-art methods}
        \label{tab:qual_comp}
        \begin{tabular}{p{1.9cm}p{0.5cm}p{0.6cm}p{0.7cm}p{0.7cm}p{0.7cm}p{0.7cm}}
        \hline
        \multirow{2}{*}{\textbf{Method}} & \textbf{Task Planning} & \textbf{Motion Planning} & \textbf{Adapt-able} & \textbf{Explain-able} & \textbf{Inter-active} & \textbf{Learn-able} \\
        \hline        
                
        \textbf{ReAct} \cite{yao2022react} & \textbf{Yes} & No & \textbf{Yes} & Partial & \textbf{Yes} & Partial \\

        \textbf{Inner \newline Monologue} \cite{huang2023inner} & \multirow{2}{*}{\textbf{Yes}} & \multirow{2}{*}{No} &  \multirow{2}{*}{\textbf{Yes}} &  \multirow{2}{*}{\textbf{Yes}} &  \multirow{2}{*}{\textbf{Yes}} &  \multirow{2}{*}{Partial} \\
        \textbf{VoxPoser} \cite{huang2023voxposer} & \textbf{Yes} & Partial & No & No & \textbf{Yes} & Partial \\
        \textbf{GPT-4} \cite{kwon2023language}  & \textbf{Yes} & Partial & \textbf{Yes} & Limited & Limited & Partial \\
        \textbf{SayTap} \cite{tang2023saytap} & No & \textbf{Yes} & Partial & \textbf{Yes} & No & No \\
        \textbf{ChatGPT} \cite{vemprala2024chatgpt} & \textbf{Yes} & No & \textbf{Yes} & Partial & \textbf{Yes} & Partial \\
        \textbf{Ours} & \textbf{Yes} & \textbf{Yes} & \textbf{Yes} & \textbf{Yes} & \textbf{Yes} & \textbf{Yes}  \\
        \hline
        \end{tabular}
    \end{table}
%-------------------------------------------------------------------------------
\\ \textbf{Low-Level Control} \cite{tang2023saytap} use foot contact pattern as a communication interface to transform human language commands into low-level quadruped robot control. They define a prompt structure that specifies the role of the LLM, the movement descriptions, general pattern rules, and finally some examples for few-shot response fine-tuning. The control pattern output is then transferred to a locomotion controller. The pattern generation is generalized enough also to understand indirect commands and infer the robot's behavior. \cite{kwon2023language} experiment on the use of LLMs in generating low-level control without providing control examples for few-shot fine-tuning. 
\\ \textbf{LLMs as General Pattern Machines:} \cite{mirchandani2023large} showcases the capability of LLMs to utilize arbitrary symbol sequences as input patterns for interpretation, synthesis, and application in real-world tasks. The examples include (i) sequence transformation: where a symbolic representation of a pattern is synthesized for transformations and validated on the abstract reasoning corpus \cite{chollet2019measure}. (ii) Sequence completion: where a numeric representation of simple function classes is extrapolated. This was extended to robotic tasks such as drawing or sweeping, where the human demonstrates an initial trajectory which later is extrapolated through LLM. (iii) Sequence improvement: includes online policy tuning involving symbolic and numeric patterns. It was demonstrated on the CartPole \cite{towers_gymnasium_2023} control optimization problem and teaching robot via clicker signals. 
\subsection{Contributions} This paper investigates the capabilities of LLMs and VLMs in interpreting human language for low-level control, generating complex trajectories, and acquiring contextual awareness of the environment.The study shows that these models can independently make informed decisions, interpret indirect communication cues, and sustain situational awareness by analyzing images that represent the robot and its environmental conditions. Furthermore, we show that information on the robot's physical structure results in an awareness of physical limitations and safe command execution. Our experimental results are directed towards developing interpretable, safe, and human-centric robotic systems, a concept which is validated via simulated robotic arm manipulation and a human-robot communication interface. Table \ref{tab:qual_comp} presents a qualitative comparison of our proposed research concepts against the current state-of-the-art.
%===============================================================================
%===============================================================================
\section{Methodology}
\label{sec:method}
In this section, we define the methodology used for addressing the research questions of LLM-based robot control and feedback on robot states. Firstly, we discuss the communication framework to bridge human language with control and perception. Secondly, we define the movement descriptions and general pattern rules used for the translation of human command to low-level control. Lastly, we define the prompt structures used in this study.

\subsection{Framework}
\label{sec:methodA}
\begin{figure}[h!]
   \centering
    \centerline{\includesvg[inkscapelatex=false, width=\columnwidth]{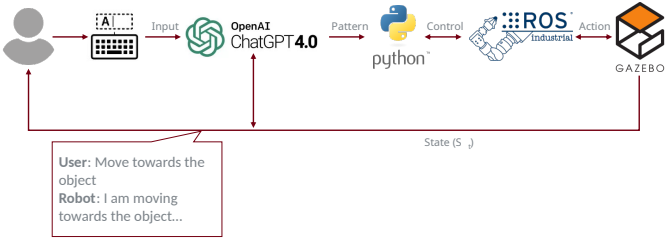}}
    \caption{Proposed framework: human-robot interaction interface from language to command and visual environment perception to human language.}
    \label{fig:frame}
\end{figure}

The framework designed for evaluating LLMs in robotic manipulation and perception is shown in Fig. \ref{fig:frame}. The interface allows the user to input a text and/or image prompt which is passed to GPT-4 through a Python client developed by \cite{vsakkas2024sydney}. The output from GPT is parsed through Python and the control commands from the response are extracted. The control commands are then transferred via ROS industrial \cite{ros-industrial2024} controller to a Gazebo simulation environment \cite{gazebosim2024gazebo}, chosen for its ease of simulation-to-reality transfer. For environment perception, the visual and/or textual observations from the simulation are fed back to GPT-4 to generate a response in human language. In our workflow LLM can play a dual role: interpretation for control and for perception. This modular approach can help to run both in parallel or other cases use the conventional approach for control and the perception module of LLM as a safety check. 
%-------------- FIGURE 1 --------------------------------------
%-------------------------------------------------------------
\subsection{Movement Descriptions and General Pattern Rules}
\begin{figure}[h!]
        \centering
        \centerline{\includesvg[inkscapelatex=false, width=0.25\columnwidth]{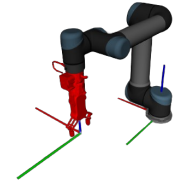}}
        \caption{Illustration of the axes of robot movement.}
        \label{fig:movement_desc}
    \end{figure}

Robot movements are described along the $X$, $Y$, and $Z$ axes, representing left/right, forward/backward, and up/down directions respectively, with movements quantified as multiples of 0 (no movement) or ±1mm. The movement patterns are formatted in $X$, $Y$, $Z$, and $G$, where $G$ represents the binary gripper control, with only three possible values for each axis: -1, 0, and 1, indicating negative movement, no movement, and positive movement. The gripper control values are 0 and 1, indicating open and closed states. Each pattern comprises four labeled lines: $X$ for left/right, $Y$ for forward/backward, $Z$ for up/down, and $G$ for gripper state, where 1 signifies a closed gripper to grasp objects and 0 signifies an open gripper to release objects. The robot's movement and gripper state can be represented by the vector $\mathbf{M}$ as shown in Equation~\ref{eq:pattern} that defines the robot's movements along the $X$, $Y$, and $Z$ axes (see \ref{fig:movement_desc} for illustration, where $x$, $y$, and $z$ are represented by red, green, and blue color, respectively) as well as the gripper state. 
%--- robot movement space ---------------------
    \begin{equation}
        \mathbf{M} = \begin{pmatrix}
        X \\
        Y \\
        Z \\
        G
        \end{pmatrix}
        \label{eq:pattern}
    \end{equation}
    where:
    \vspace{-0.25cm}
    \begin{align*}
        X & \in \{-1, 0, 1\} & \text{\small{(Left/Right movement: -1mm, 0mm, +1mm)}} \\
        Y & \in \{-1, 0, 1\} & \text{\small{(Fwd/Bwd movement: -1mm, 0mm, +1mm)}}\\
        Z & \in \{-1, 0, 1\} & \text{\small{(Up/Down movement: -1mm, 0mm, +1mm)}} \\
        G & \in \{0, 1\} & \text{\small{(Gripper state: 0 = open, 1 = closed)}}
    \end{align*}
    \vspace{-0.8cm}
%-----------------------------------------------    
\subsection{Prompt Structures}
The proposed methodology involves the development of concepts for language-based control and verbalized machine states. These solutions aim to improve the communication between humans and robots, enhancing the human operator experience. We aim to bridge the communication gap between human language to low-level robot control for achieving tasks such as grasping the object, moving the object, placing the object, obstacle avoidance, and further typical tasks. LLM-assisted control and task interpretation was examined in a gradual manner (see Fig. \ref{fig:prompt_struct}), where incrementally added information was introduced (as part of the LLM-input) to facilitate LLM-based reasoning and control. In the followings, we describe these input information increments in more detail.

  \subsubsection{Baseline Control Prompt Structure}
  \label{para:base_control}
\begin{figure}[t!]
    \centering
    \includegraphics[width=1.0\columnwidth]{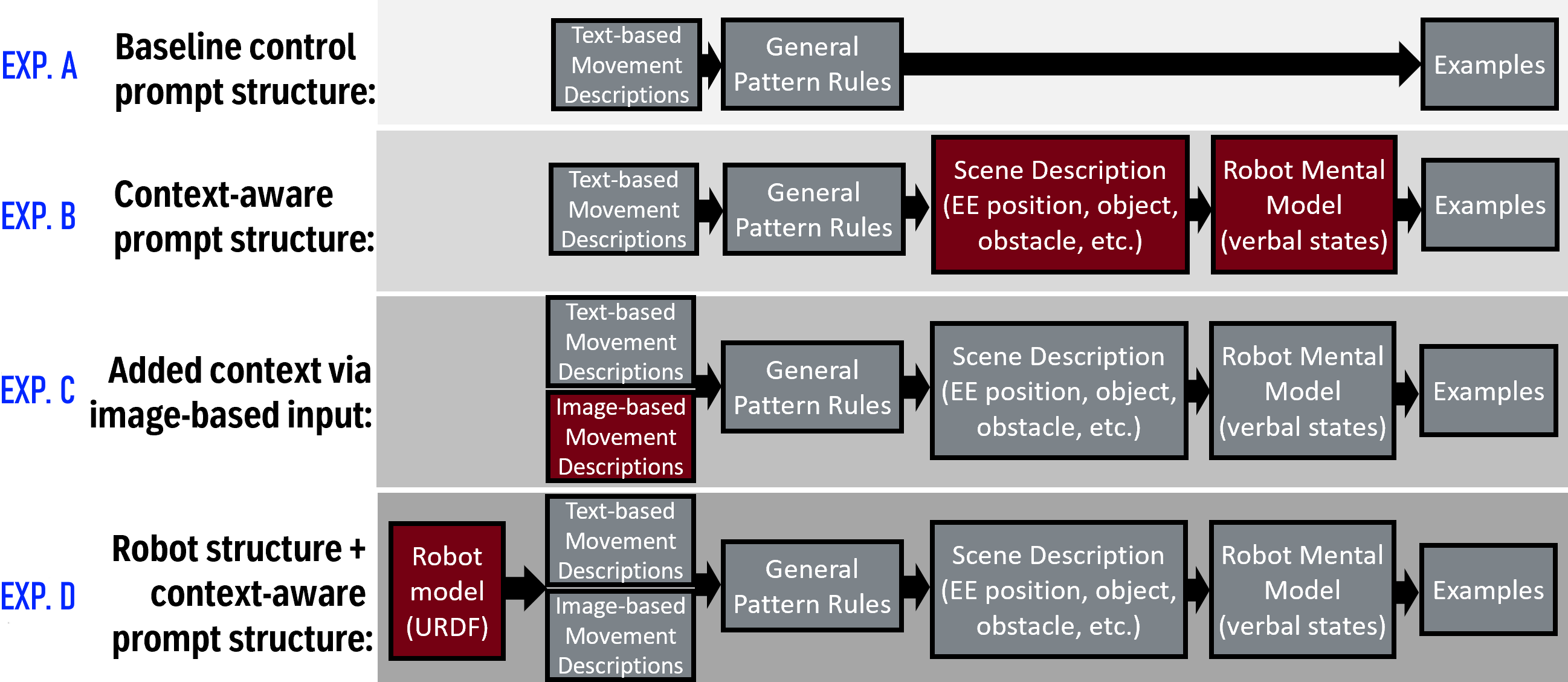}
    \caption{Prompt structures with incrementally added information, facilitating LLM-based reasoning and robot control. Blue indices refer to the experiments in Section \ref{sec:results}.}
    \label{fig:prompt_struct}
\end{figure}

  The baseline prompt structure is inspired by the prompt structure and pattern rule defined by \cite{tang2023saytap}. It starts by describing the role of the LLM, followed by basic movement definitions and descriptions. Once the movement primitives are defined, a specific task is defined using the pattern rules defined before, in form of a human language input. The outputs are then concretely defined by a few examples for fine-tuning (few-shot prompting practice). The structure is shown in the top row of Fig. \ref{fig:prompt_struct}. %As an additional perception layer, the same output by the LLM can be fed back to the LLM to minimally describe the desired action of the robot to the operator to enhance interpretability.
%--------------- FIGURE 3 ----------------------------------------------------            

\subsubsection{Context-Aware Perception: Verbalized Current Machine States}
\label{subsubsection:perc_control}
    Apart from controlling the robot through human language, there is another key objective in safety-critical applications: to perceive the environment and plan/execute machine actions accordingly. Therefore, we propose to use LLMs/VLMs to (i) capture the states of the machine and the environment from text and/or image, as \textit{perception}, and (ii) to communicate these states to the operator. Thus, in a second set of experiments, we extend the baseline control prompt (Section \ref{para:base_control}) to provide information on internal (\textit{EE} = end-effector) and external (scene/object) states. The additional information on internal and external states injected into the LLM results in a more complete awareness about its internal states and surrounding context (denoted as its internal \textit{mental model}). Turning this internal model into text, or the verbalization of robot states represents thus in this workflow an additional key added functionality. Such a verbalization shall render machine/robot operation more transparent towards a user. This extended set of prompt-based information is depicted in the second row of Fig. \ref{fig:prompt_struct}.
          
    \subsubsection{Added Perception Through Visual Information}
    The visualization of the robot simulation (see Section \ref{sec:methodA}) can be rendered into one or multiple images, depicting the robot pose/state from an external observer's point-of-view. Thus these images can enhance the prompt in form of a \textit{vision-based perception} through one or a stack of visual frames. Thus image-based input can complement the previously described text-based movement input. Furthermore, this additional input can be added in a time-incremental fashion, where image frames and text-based state observations are captured per time-step and then are transferred to LLM/VLM after a partial task completion or after a pre-defined step size. To enhance robustness, the image frame stacking step was extended to include image frames from multiple viewpoints. These frames are stacked together along the horizontal dimension, forming image-sets organized in image rows. This added set of input modality is shown in the third row of Fig. \ref{fig:prompt_struct}.
    
    \subsubsection{Enhancing Awareness to Robot Structure}
    Information on the physical structure of the robot can markedly complement the LLM's awareness to internal states and its degrees of freedom. This integration allows the LLM to generate more accurate and executable task plans by considering the robot's physical capabilities and limitations. Integrating information about the physical structure can be provided in two ways: (i) by a URDF model (interpreted as text input) or any other representation (e.g. technical specification sheet); and/or (iii) through visual means depicting the robot's physical structure. 
        
    %\subsubsection{Prompt Engineering}
    %The human-designed prompt may not be optimal for how the LLM learns to generate responses. For that reason, the prompts were optimized through prompt engineering (see \ref{app:prompt_engr_claude,app:prompt_engr_gpt}) and prompt combination (see \ref{app:comb_prompt}).

%===============================================================================
\section{Design of Experiments}
\label{sec:doe}
The design of experiments is split into two main categories: (i) LLM-based control and (ii) LLM/VLM-based perception, which focuses on situational and structural awareness.

    \subsection{Language-Based Control}
\begin{figure}[h!]
        \centering
        \centerline{\includesvg[inkscapelatex=false, width=0.7\columnwidth]{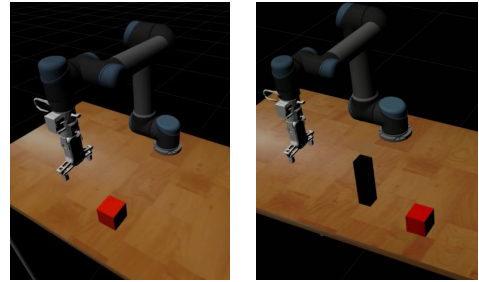}}
        \caption{Design of experiments for control.}
        \label{fig:doe}
    \end{figure}

    The scenarios used for validating language-based control involve (i) grasping task and (ii) grasping task with obstacle avoidance. Fig. \ref{fig:doe} shows the corresponding setup, consisting a red object and a black obstacle. Initial observations in form of object type and positions are provided as prompts before processing the task.

    \subsection{Perception}
\begin{figure}[h!]
        \centering
        \includegraphics[width=0.7\columnwidth]{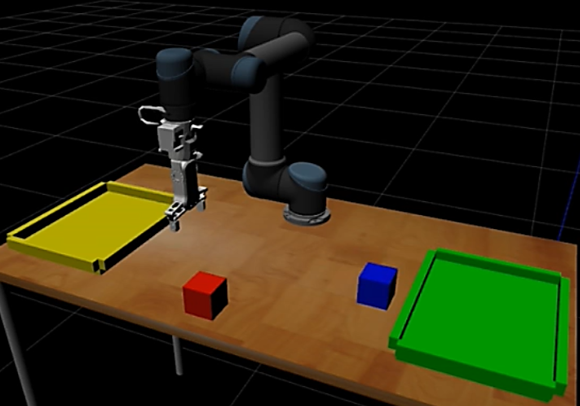}
        \caption{Design of experiments for perception through LLM.}
        \label{fig:doe_perc}
    \end{figure}

    Perception is validated in terms of both control and safe operation. The tasks performed include (i) a pick-and-place task and (ii) a grasping task that requires obstacle avoidance. The perception involves validating the consequences of actions, interpreting the environment, and placements of the objects in the environment. We add the possibility to assign real-world object and enviroment properties within the scene, as shown in Fig. \ref{fig:doe_perc}. Different colors in the figure imply different attributes, which can be defined in the prompt. For raising awareness w.r.t.  robot's and environment' physical structure, we provide the robot URDF \cite{tola2024understanding} description (text from an \textit{xml} file) and textual description of the environment. The prompts are provided with the initial observation list before progressing with the task.
    
\begin{figure}[h!]
        \centering
        \includegraphics[width=\columnwidth]{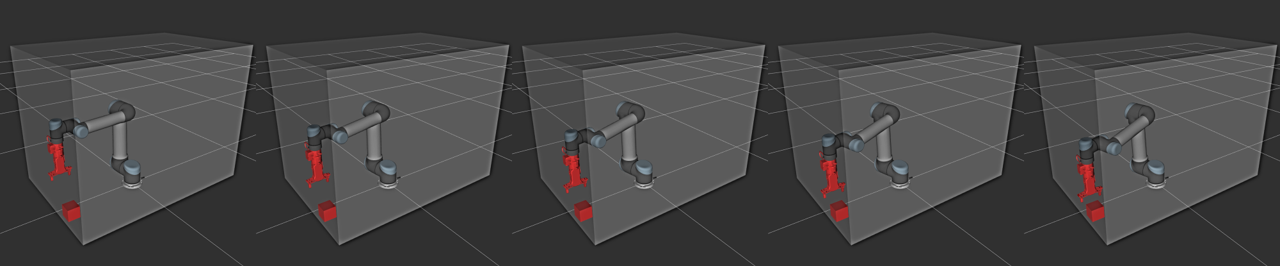}
        \caption{Single-view evolution of an executed task within a safe zone, generated as a frame stack.}
        \label{fig:doe_perc_frame}
    \end{figure}

\begin{figure}[h!]
        \centering
        \includegraphics[width=\columnwidth]{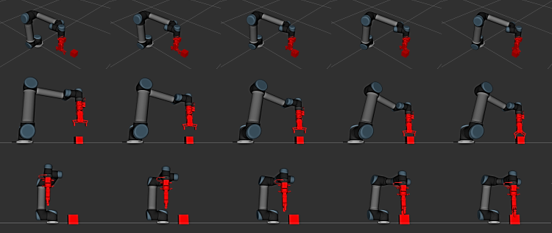}
        \caption{Multi-view evolution of a grasping task, generated as a frame stack.}
        \label{fig:doe_perc_frame_multi}
    \end{figure}

    Furthermore, to enhance the capabilities of perception, we use VLM to ingest a stack of image frames, which are time-incremental rendered views from external (with respect to the robot) viewpoints. Multiple views, highlighted robot/object parts and volumes (e.g. safe zone) are used in the experiments. Examples are shown for a single view (Fig. \ref{fig:doe_perc_frame}) and for a multi-view case (Fig. \ref{fig:doe_perc_frame_multi}). The figures show a scenario where robot manipulation involves a red object, whereas the gray bounding box representing a safe operational zone. Highlighted objects and spatial relations are on one hand to facilitate VLM in its reasoning task, on the other hand it aids the user in understanding the time-evolution of executed actions.
   
%===============================================================================
\section{Experiments}
\label{sec:results}
In this section, we address the research questions of Section \ref{sec:intro} as experiments. Experiments follow the structure depicted in Fig. \ref{fig:prompt_struct}.

\subsection{Baseline Language-Based Control}
These experiments involve (i) control pattern optimization and  (ii) text-based perceptive control by providing initial observations of the environment.
    
\subsubsection{Baseline Control Pattern}
For the low-level control, first, we employed a control pattern similar to \cite{tang2023saytap} and adapted it for robot manipulation. Full prompt of the pattern rule definitions is given in Appendix \ref{app:base_prompt}.

\begin{lstlisting}
Input: Move forward and pick the cube 
Output:     X: 00000000000000000000000000 
            Y: 11111111111111111111111111 
            Z: 00000000000000000000000000 
            G: 00000000000000000000000001
            
Input: Move backward and release the cube 
Output: 
    X: 00000000000000000000000000 
    Y:-1-1-1-1-1-1-1-1-1-1-1-1-1-1-1-1-1-1-1-1-1-1-1-1-1-1 
    Z: 00000000000000000000000000 
    G: 11111111111111111111111110

Input: Move right 
Output:     X: 11111111111111111111111111 
            Y: 00000000000000000000000000 
            Z: 00000000000000000000000000
            G: 00000000000000000000000000
\end{lstlisting}

The baseline control experiments exclude environmental observations and focus on evaluating the prompt structure and control patterns. They involved general movement control, grasping tasks, and grasping while avoiding obstacles, with initial tests conducted without gripper actions.

\subsubsection{Improved Control Pattern}
There were several drawbacks involved with the baseline control pattern due to its highly redundant content: (i) duration to generate the prompt, (ii) lower accuracy (\textit{cm} accuracy), (iii) higher failure rate due to repetitive pattern generation and loss of information in memory. To improve pattern execution, two changes were implemented: (i) modifying the control pattern structure to include multiples of 0s, 1s, or -1s, and (ii) breaking the trajectory into steps to improve task robustness, as demonstrated in an ablation study by \cite{kwon2023language}. The improved control pattern is shown below:

\begin{lstlisting}
Input: Move forward 100mm and pick a cube
Output:     X: [0]*100
            Y: [1]*100
            Z: [0]*100
            G: [0]*99 + [1]*1

Input: Move backward 50mm and release the grasped cube
Output:     X: [0]*50 
            Y: [-1]*50 
            Z: [0]*50 
            G: [1]*49 + [0]*1

Input: Move right for 70mm
Output:     X: [1]*70 
            Y: [0]*70 
            Z: [0]*70 
            G: [0]*70
\end{lstlisting}
This improvement helped in (i) faster response, (ii) higher accuracy (\textit{mm} accuracy), (ii) lower failure rate, and (iv) improved generalization. The comparison for improved generalization is shown below which shows the control pattern generalizing to complex functional representation rather than just 0s and 1s:
\begin{lstlisting}
Experiment: Circular movement
Input: Move in a circle
Output:     X: 111110000-1-1-1-1-1-1-1-100001111 
            Y: 000111111110000-1-1-1-1-1-1-1-100 
            Z: 0000000000000000000000000 
            G: 0000000000000000000000000
    becomes:
Input: Move in a circle
Output:     X: [cos(t) for t in range(360)]
            Y: [sin(t) for t in range(360)]
            Z: [0]*360
            G: [0]*360
\end{lstlisting}

\begin{lstlisting}
Experiment: Obstacle avoidance
Input: Move in a pattern to avoid an obstacle in a path and pick the cube on the right
Output:     X: [0]*10 + [1]*50 + [0]*10 + [0]*1 
            Y: [0]*71 
            Z: [1]*10 + [0]*50 + [-1]*10 + [0]*1 
            G: [0]*70 + [1]*1
\end{lstlisting}

\subsubsection{Comparative Analysis}
We compare the improved control pattern strategy with the baseline adapted from SayTap \cite{tang2023saytap}. A comparison is shown in Table \ref{tab:quan_comp} for tasks involving grasping and obstacle avoidance. The average time for pattern generation, final goal error, and success rate of completing the task were calculated over 10 runs on the prompt for each task with variations in the prompt sentence structure and morphology, grasping object, end effector, and obstacle positions. The test prompts for both methods were consistent. From the results of obstacle avoidance, it can be observed that TalkWithMachines is slower due to the generation of comparatively complex trajectories, however, has a better spatial accuracy and success rate.
\begin{table}[ht]
    \centering
    \caption{Comparison with the prior art's control pattern generation strategy.}
    \label{tab:quan_comp}
    \begin{tabular}{p{1.3cm}p{0.7cm}p{0.7cm}p{0.7cm}p{0.8cm}p{0.8cm}p{0.8cm}}
        \hline
        \multirow{2}{*}{\textbf{Task}} & \multicolumn{3}{c}{\textbf{SayTap} \cite{tang2023saytap}} & \multicolumn{3}{c}{\textbf{Ours}}                        \\ \cline{2-7} 
                                       & \textbf{Time (s)}             & \textbf{Error (m)}             & \textbf{Success}             & \textbf{Time (s)} & \textbf{Error (m)} & \textbf{Success} \\ \hline
        Grasping                 & 6.90                              & 0.01                       & 0.90                          & \textbf{6.10}                   & \textbf{0.00}           & \textbf{1.00}              \\
        Obstacle avoidance             & \textbf{5.84}                               & 0.05                       & 0.3                          & 6.82                  & \textbf{0.018}           & \textbf{0.5}              \\
        \hline
    \end{tabular}
\end{table}

\subsection{Context Aware Reasoning and Verbalized Machine States}
These experiments emphasize text-based perception prior to or during manipulation. Perception enables LLMs to ask users for clarification if commands are incomplete or if the environment context needs further details (e.g., unobserved objects to grasp or dangerous action). Furthermore, the LLM is able summarize the planned action in a step-wise manner and indicate the validity of each step.
    
\begin{lstlisting}
Experiment: Obstructed position avoidance
Input: place the grasped red cube to position  [-270, 190, 30] with observation = [[110, 490, 140], [-170, 190, 30], [-270, 190, 30]]
Output: Given the current position of the end effector with the grasped red cube at [-170, 190, 30], and the desired location to place the red cube at [-270, 190, 30], the end effector needs to move from its current position to the desired location. However, there is a blue cube already present at the desired location [-270, 190, 30]. Placing the red cube at this location would result in the red cube penetrating the blue cube, which is not possible as the cubes are solid and impenetrable. Therefore, this action is not possible. Please provide a different location to place the red cube. Let me know if you need help with anything else.
\end{lstlisting}
LLM can also reason about possible outcomes based on object and scene real-world attributes:
\begin{lstlisting}
Experiment: Awareness to real-world attributes
Input: Move the red cube to the yellow zone
Output: Moving the red cube to the yellow zone would be dangerous as the red cube is made of wood and the yellow zone represents fire. This could potentially start a fire. Therefore, this action is not recommended. Here's the explanation:
The red cube is made of wood, which is a combustible material. 
The yellow zone represents fire, which can ignite combustible materials. 
Moving the red cube to the yellow zone could potentially start a fire, causing a dangerous situation. 
So, I would advise against performing this action. Safety should always be the top priority when operating robotic systems. Please provide another task that doesn't involve potential hazards.

Input: Move the blue cube to the yellow zone
Output: Given the environmental constraints and the properties of the objects, moving the blue cube (made of ice) to the yellow zone (representing fire) would be dangerous as it could cause the ice to melt. Therefore, I cannot provide a control pattern for this action. It's crucial to ensure the safety of the environment and respect the real-world properties of the objects. Please provide another task that doesn't violate these properties.
\end{lstlisting}

\subsection{Added Context via Image-based Inputs}
In these experiments, we test VLMs for becoming aware of spatial relations within the environment and detecting anomalies. First, we experiment with a single visual stacks of frames as shown in Fig. \ref{fig:cube_coll_vlm_sing}. The full prompt of the experiment is given in Appendix \ref{app:perc_sing}.

\begin{figure}[h!]
            \centering
            \includegraphics[width=\columnwidth]{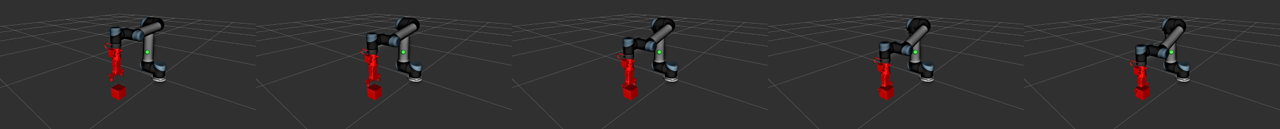}
            \vspace{-0.75cm}
            \begin{lstlisting}
Output: Collision imminent.
Reason: End effector descends, force negative. 
            \end{lstlisting}  
            \caption{Perception via image-based observations and text descriptions.}
            \label{fig:cube_coll_vlm_sing}
        \end{figure}  

\textbf{Region of Interest for Enhanced Accuracy}
When providing an enlarged view of a specific region of interest, it helps to identify a failure case, whereas a far-away view does not suffice to solve the task.
        \begin{figure}[h!]
            \centering
            \includegraphics[width=\columnwidth]
            {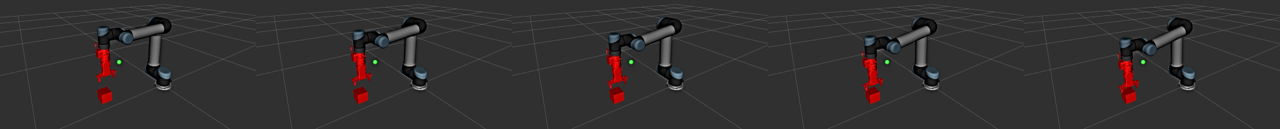}
            \vspace{-0.75cm}
            \begin{lstlisting}
Output: **No collision, safe grasp.**
Reason: **Cube aligns with gripper path.**
            \end{lstlisting}  
            \includegraphics[width=\columnwidth]
            {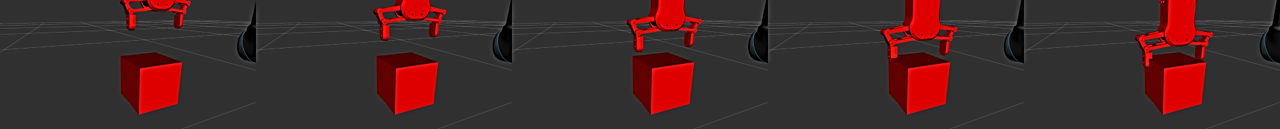}
            \vspace{-0.75cm}
            \begin{lstlisting}
Output: Collision imminent.
Reason: Cube misaligned, end effector close.  
            \end{lstlisting}  
            \caption{Perception with far-away (top) and enlarged view (bottom) of a region of interest.}
            \label{fig:roi}
        \end{figure}  

\subsection{Robot Structure Awareness via URDF import}
We added structural perception via URDF (Universal Robotic Description Format) (see Fig. \ref{fig:prompt_struct}), which involves defining each robot component, its shape and limitations and permits the LLM to acquire a structural concept about the robotic arm. To verify the acquired percept within the LLM, a visualization of the hierarchical robot structure was generated using an LLM-generated \textit{Mermaid} \cite{mermaid-js} code of the robot part hierarchy. The generated 10-level hierarchical visualization correctly captured the robot's physical structure (not shown).

\subsection{Experiments using the complete workflow D}
Using the workflow \textit{D} (Fig. \ref{fig:prompt_struct}) including image-, text- and URDF-based information on external scene and robot-internal structure, we carried out additional experiments.
\subsubsection{Operation within a Safe Zone}
\begin{figure}[h!]
                \begin{center}
                \includegraphics[width=.8\columnwidth]
                {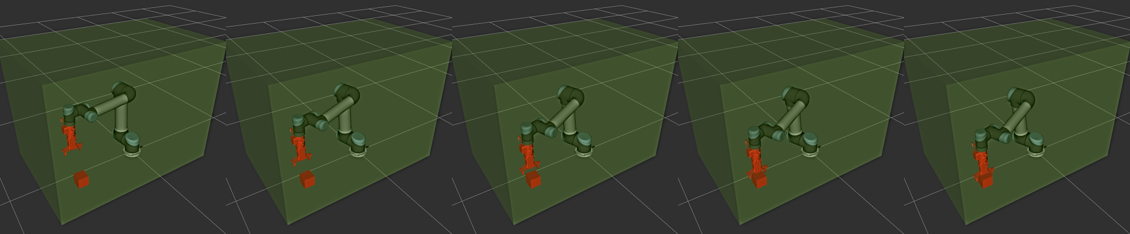}
                \vspace{-0.3cm}
                \begin{lstlisting}
Output:     No collision; grasp in range.
            Red cube clear; path safe.
                \end{lstlisting}  
                \includegraphics[width=.8\columnwidth]
                {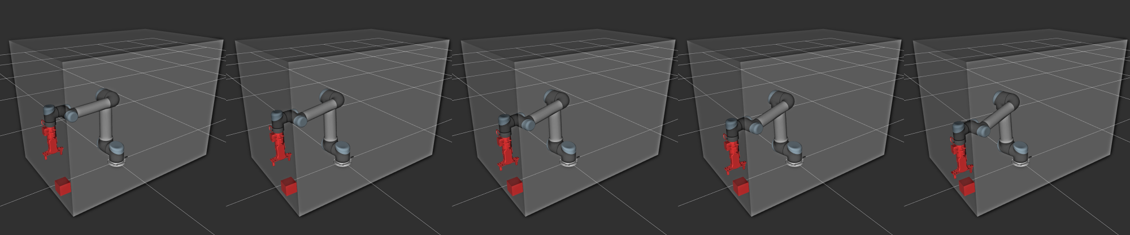}
                \vspace{-0.3cm}
                \begin{lstlisting}
Output:     Collision imminent.
            Red cube misaligned; wall proximity.
                \end{lstlisting}  
                \caption{Spatially-aware task executions via text (scene+URDF) and visual inputs, using a single-viewpoint frame stack.}
                \label{fig:struct_perc_sing}
                \end{center}
\end{figure}

A cuboid-shaped safe operation zone was defined by text and image-stack prompts, as an environment constraint. The correct VLM responses in two scenarios are shown in Fig. \ref{fig:struct_perc_sing}. 

\subsubsection{Obstacle Avoidance}
\begin{figure}[h!]
            \centering
            \begin{lstlisting}
Input: Move towards the object. <observation list>
            \end{lstlisting}
            \vspace{-0.3cm}
            \includegraphics[width=\columnwidth]{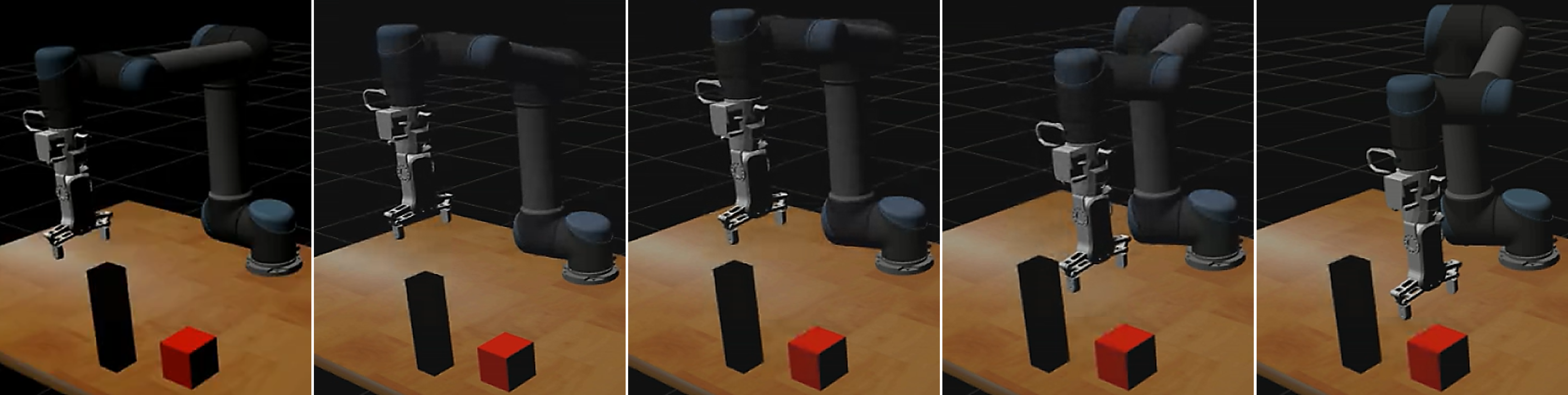}
            \caption{Result image sequence showing a successful end effector approach while avoiding an obstacle (black object).}
            \label{fig:obst_task}
        \end{figure}

Fig. \ref{fig:obst_task} illustrates the task execution for obstacle avoidance using human language and URDF input only. It can be seen that the LLM can correctly infer the necessary obstacle avoidance from the input observation list, even without explicit related instructions.
\subsubsection{Stacking Operation}
\begin{figure}[h!]
            \centering
            \begin{lstlisting}
Input: Move the red cube on top of the blue cube. <observation list>
            \end{lstlisting}
            \vspace{-0.3cm}
            \includegraphics[width=\columnwidth]{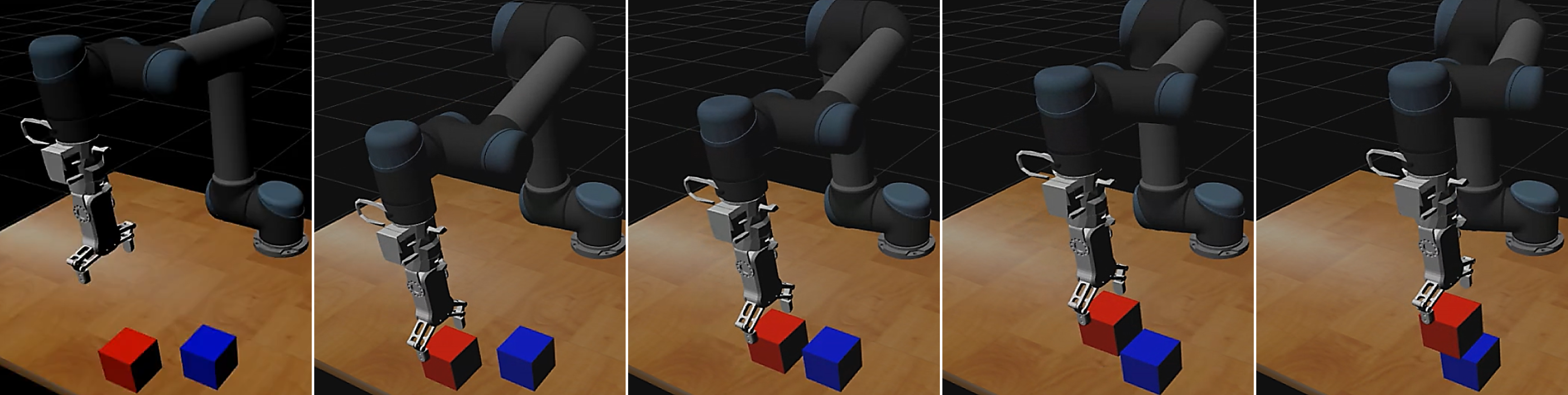}
            \caption{Result of a stacking task, shown as a generated image sequence.}
            \label{fig:stack_task}
\end{figure}

In this successful task demonstration (see Fig. \ref{fig:stack_task}), multiple spatial reasoning steps are involved: reaching and grasping the object, moving it to a safe height, and placing it on top of another object based on the positions and dimensions listed in the input observations.
        \subsubsection{Pick and Place Into a Zone}
\begin{figure}[h!]
            \centering
            \begin{lstlisting}
Input: Grasp the red cube and place it in zone A. <observation list>
            \end{lstlisting}
            \vspace{-0.3cm}
            \includegraphics[width=\columnwidth]{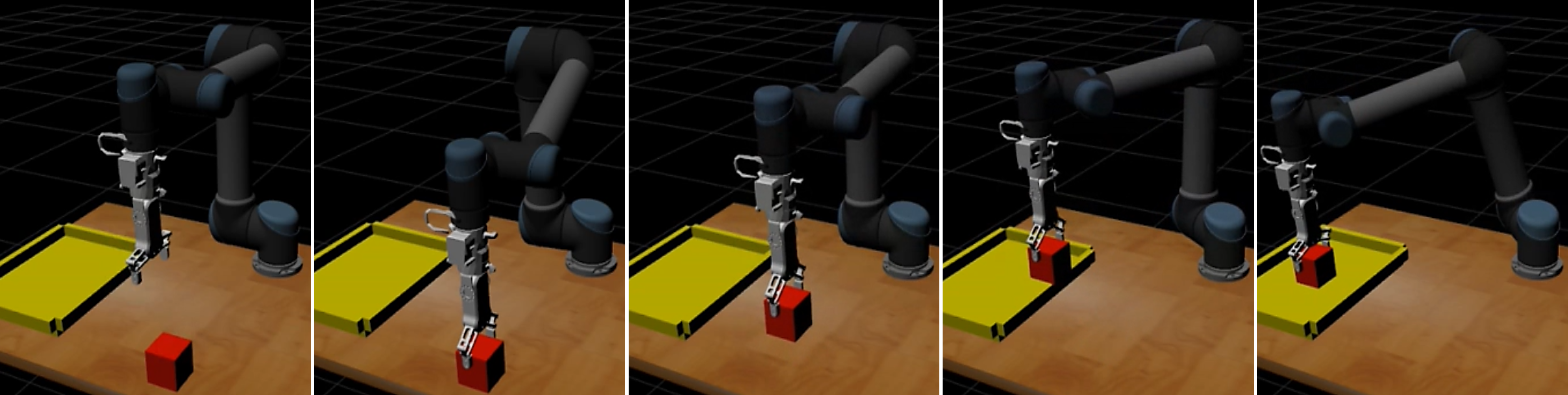}
            \caption{Result of the task execution moving an object into a zone.}
            \label{fig:zone_task}
        \end{figure}

        The successful task execution shown in Fig. \ref{fig:zone_task} involves placing an object into a user-defined zone, with the necessary information provided in the initial observation list.
\subsubsection{Attribute-based Object Sorting into Respective Zones}
\label{subsec:sort}
In this example control pattern generation considers real-world object/scene attributes to generate sorting decisions. Such a workflow imitates an industrial sorting scenario with complicated decision logic. An LLM is equipped with a foundation-level world knowledge, hence complex logic rules do not need to be explicitly defined. 
\begin{lstlisting}
Input: Move the cubes to the zones
Output: Given the physical properties of the cubes and the zones, it would be hazardous to move the red cube (Fire) to the green zone (Forest) as it could potentially cause a fire. Similarly, moving the blue cube (Ice) to the yellow zone (Fire) would not be appropriate as the ice would melt. Therefore, I will provide a control pattern that safely moves the red cube to the yellow zone and the blue cube to the green zone.
Control Pattern for Red Cube (Fire) to Yellow Zone (Fire):
  generated control sequence (not shown in detail)
Control Pattern for Blue Cube (Ice) to Green Zone (Forest):
  generated control sequence (not shown in detail)
\end{lstlisting}
\textbf{Observed failure modes:} From all conducted experiments, it can be concluded that tasks involving the spatial interaction of a small number of objects (such as the end effector, a single target, and one obstacle) were consistently successful. However, failures occurred in spatial reasoning when numerous closely spaced objects or strict spatial constraints (such as a tight safety zone) needed to be taken into account during the task. Enhancing spatial resolution (for example, through a zoomed-in view as shown in Fig. \ref{fig:roi}) helped to mitigate some of these failure modes.
%===============================================================================
\section{Conclusions and Future Work}
\label{sec:conclusion_future}
This paper illustrates the integration of large language models and vision-language models to improve human-robot interaction. Through experiments involving robotic arm manipulation in simulated environments, we demonstrated that LLMs can successfully generate low-level control commands, perceive their surroundings, and communicate effectively with humans. Notable findings include the capability of LLMs to create intricate trajectories, comprehend environmental constraints, and make autonomous decisions. Future work will focus on implementing these systems in real-world scenarios, enhancing contextual and structural understanding through real-time visual streams, developing self-learning mechanisms, and exploring applications in safety-critical industries. The goal is to create interpretable and safe robotic systems that enhance user trust and foster natural human-robot interaction.

\bibliography{talk_machines}
%===============================================================================
\newpage
\appendix
\section{Control Prompts}
\subsection{Baseline Pattern Rule}
\label{app:base_prompt}
\begin{lstlisting}
# Role
You are a robot control pattern manipulation expert.
Your job is to give an end effector position control pattern in the format of multiples as will be shown in the examples at the end based on the input.
Assume the necessary conditions.
The robot moves simultaneously in the directions defined.
You will always give the output in the correct format no matter what the input is.
Just give the control pattern and avoid too much explanation.

# Movement Descriptions
The following are descriptions of robot movements:
 1. Moving left or right is represented as moving in the positive or negative X direction for 1mm or -1mm, respectively.
 2. Moving forward or backward is represented as moving in the positive or negative Y direction for 1mm or -1mm, respectively.
 3. Moving up or down is represented as moving in the positive or negative Z direction for 1mm or -1mm, respectively.

# General Pattern Rules
 The following are rules for describing the robot movement patterns:
 1. You should output the movement patterns in X, Y, and Z format and the gripper binary control in G format.
 2. There are only three values to choose from for each of the axes: [-1, 0, 1], which represents movement along that axis.
 3. There are only two values to choose from for gripper control [0, 1], which represents the gripper closed or open.
 4. A pattern has four lines, each of which represents the robot movement pattern of the end effector and gripper control.
 5. Each line has a label. "X" for the movement in the left or right direction, "Y" for the movement in the forward or backward direction, and "Z" for the movement in the up or down direction. "G" represents gripper open or close.
 6. For the first three lines (X, Y, and Z), "0" represents no movement in that direction, "1" represents positive movement in that direction for 1mm, and "-1" represents negative movement in that direction for -1mm. For the fourth line (G), "0" represents the gripper opened, and "1" represents the gripper closed. If the object has to remain grasped, the gripper control should be 1 and to release the object the gripper value should be 0.

# Examples
Input: Move forward 100mm and pick a cube
 X: [0]*50
 Y: [1]*100
 Z: [0]*30
 G: [0]*99 + [1]*1

Input: Move backward 50mm and release the grasped cube
 X: [0]*10
 Y: [-1]*50
 Z: [0]*20
 G: [1]*49 + [0]*1

Input: Move left for 70mm
 X: [1]*70
 Y:  [0]*20
 Z: [0]*10
 G: [0]*70
\end{lstlisting}

\newpage
\hfill
\vspace{-0.1cm}
\subsection{Single-View Image Stack with Observation Sequences}
\label{app:perc_sing}
\begin{lstlisting}
# Role
You are the robot and the sensor observation is given in a list of observations.

# Observation Description
Each observation list is ordered as:
                            [[if the cube is grasped or not in a boolean value],
                             [position of end effector (x, y, z)],
                             [velocity of end effector (x, y, z)],
                             [red cube position (x, y, z)],
                             [blue cube position (x, y, z)],
                             [force on end effector in z]]
To help visually, the camera image is given as a real-time frame stack starting from left. 

# Task Description
It is a grasping task with the object being the red cube. 

# Constraints
The object should be properly aligned in the gap with the gripper fingers otherwise the object will collide with it while the gripper is moving toward it. 

# Objective
Describe the robot's state and if at any point it is going to or has already collided etc.

# Safety Checks 
Also, check if the black cube does not obstruct the red cube. Predict the future state or if any dangerous anomaly is about to occur.

# Output
give output response only in 50 characters.
give the reason for the decision based on observation sequence or images only in another 50 characters.
\end{lstlisting}
Input image stack:
\begin{figure}[h!]
    \includegraphics[width=\columnwidth]{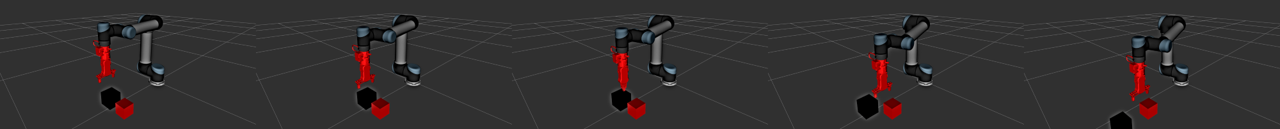}
\end{figure}

\end{document}